\documentclass{article}

 \usepackage[main, final]{neurips_2026}

\usepackage[utf8]{inputenc} 
\usepackage[T1]{fontenc}    
\usepackage{hyperref}       
\usepackage{url}            
\usepackage{booktabs}       
\usepackage{amsfonts}       
\usepackage{nicefrac}       
\usepackage{microtype}      
\usepackage{xcolor}         
\usepackage{amsfonts}
\usepackage{amsmath}
\usepackage{amssymb}
\usepackage{graphicx}
\usepackage{algorithm}
\usepackage{algorithmic}
\usepackage{multirow}
\usepackage{bm}
\usepackage{mathtools}
\usepackage{subcaption}
\usepackage{wrapfig}
\usepackage{fvextra}
\usepackage{svg}

\newcommand{\RR}{\mathbb{R}}
\newcommand{\EE}{\mathbb{E}}
\newcommand{\calF}{\mathcal{F}}
\newcommand{\calX}{\mathcal{X}}
\newcommand{\calA}{\mathcal{A}}

\newcommand{\calR}{\mathcal{R}}
\newcommand{\calP}{\mathcal{P}}

\newcommand{\bz}{\bm{z}}
\newcommand{\blam}{\bm{\lambda}}
\newcommand{\bc}{\bm{c}}
\newcommand{\balpha}{\bm{\alpha}}
\newcommand{\bbeta}{\bm{\beta}}

\newcommand{\argmax}{\mathop{\mathrm{argmax}}}
\newcommand{\training}{\mathcal{D}_{\mathrm{train}}}

\DeclareMathOperator{\IC}{IC}

\DeclareMathOperator{\Corr}{Corr}

\DeclareMathOperator{\Mean}{Mean}
\DeclareMathOperator{\Std}{Std}
\DeclareMathOperator{\Rank}{Rank}
\DeclareMathOperator{\LLM}{LLM}
\DeclareMathOperator{\LSTM}{LSTM}
\DeclareMathOperator{\FFN}{FFN}
\DeclareMathOperator{\RRE}{RRE}
\DeclareMathOperator{\PFS}{PFS}

\usepackage[capitalize,noabbrev]{cleveref}

\title{AlphaPareto: Formulaic Alpha Discovery with LLM-Guided Multi-Objective Reinforcement Learning}

\author{%
  Yingbo Zhao \\
  School of Economics\\
  Xiamen University\\
  Fujian, China\\
  \texttt{15620241152783@stu.xmu.edu.cn} \\
  \And
  Zeyu Yang \\
  Paula and Gregory Chow Institute for Studies in Economics\\
  Xiamen University\\
  Fujian, China\\
  \texttt{yangzeyu@stu.xmu.edu.cn} \\
  \And
  Zhoufan Zhu \thanks{Correspondence to: Zhoufan Zhu <tylerzzf@xmu.edu.cn>.} \\
  School of Economics \& Wang Yanan Institute for Studies in Economics\\
  Xiamen University\\
  Fujian, China\\
  \texttt{tylerzzf@xmu.edu.cn} \\
}

\begin{document}

\maketitle

\begin{abstract}

    Formulaic alpha discovery is a core challenge in quantitative trading, as identifying alphas that work well together remains difficult. Recent reinforcement learning (RL) methods formulate this task as a Markov decision process (MDP), but two important issues remain unresolved. First, as the alpha pool evolves, the reward function changes accordingly, making the MDP inherently non-stationary. Second, most existing methods optimize a single objective, typically predictive power, while ignoring other important properties of a high-quality alpha pool. Motivated by these challenges, we propose AlphaPareto, an RL method for formulaic alpha discovery. To address non-stationarity, AlphaPareto augments the state to include both the alpha under construction and the current alpha pool, and applies a large language model (LLM) to encode the pool. This design allows the agent to adapt to the evolving search environment. To overcome the limitation of single-objective reward design, AlphaPareto replaces the scalar reward with a multi-objective vector-valued reward that simultaneously captures predictive power, temporal stability, perturbation robustness, and diversity, and optimizes these objectives through a Pareto-regularized learning procedure. Empirical applications to real-world datasets show that our AlphaPareto method outperforms its competitors.
    
\end{abstract}

\section{Introduction}

In quantitative investing, predictive signals derived from historical market data, known as alphas (or alpha factors), form the foundation of automated trading strategies. Among the various forms of alphas, \emph{formulaic alphas}, which transform stock features (e.g., historical prices and volume) into signals via concise mathematical formulas, are particularly valued for their interpretability, auditability, and ease of deployment. Since a single alpha is typically unstable across dynamic markets, the final trading signal is usually a \emph{mega-alpha}, which combines a set of formulaic alphas (i.e., an \emph{alpha pool}) through an interpretable model (say, a linear model). Intuitively, the quality of this mega-alpha depends critically on whether the included alphas are \emph{synergistic}, i.e., whether the alpha pool captures diverse and complementary aspects of market dynamics.

AlphaGen \citep{alphagen} introduces a powerful paradigm for formulaic alpha discovery, which formulates the discovery process as a Markov decision process (MDP) and applies reinforcement learning (RL) to generate a synergistic alpha pool. This RL-based approach has been further advanced by \cite{alphaqcm}, \cite{alphaforge}, \cite{alphasage}, among others, yet two important issues remain unresolved, limiting the efficiency and stability of the discovery process in practice.

The first issue is \emph{non-stationarity} \citep{Erwan2019nonstationary}: the reward function of the alpha discovery MDP changes across episodes. This stems from a structural misalignment between state and reward. The reward measures the contribution of the generating alpha to the current alpha pool, and therefore depends on the pool's evolving composition. However, the state encodes only the generating alpha, and no information about the pool is included. As stronger alphas enter the pool and replace weaker ones, the same state-action trajectories can yield different rewards at different stages of alpha discovery. Most existing methods simply ignore this non-stationarity \citep{alphagen,ren2024riskminer,alphaforge,alphasage}, except \cite{alphaqcm}. So far, none of the existing methods inform the RL agent about what the current alpha pool already contains or what complementary signals it should seek next; the agent must infer this indirectly from rewards alone.

The second issue is \emph{single-objective reward design}. All existing RL-based methods optimize a single scalar reward, typically the Information Coefficient (IC) of the mega-alpha formed by the alpha pool. While IC measures average predictive power, the quality of an alpha pool is inherently multi-dimensional. For example, a pool with high IC may still exhibit large temporal fluctuation or prove fragile under market regime shifts, leading to poor out-of-sample performance. As noted by \cite{ding2025alphaeval}, other criteria, including temporal stability, perturbation robustness, and diversity, are also important for a reliable alpha pool. However, no existing method incorporates these criteria into the alpha discovery process.

To address both issues, we propose \textbf{AlphaPareto}, an RL method that integrates alpha pool semantic reasoning and Pareto-regularized multi-objective RL (MORL). Specifically, to handle non-stationarity, we redefine the state to include both the generating alpha and the evolving alpha pool. Under this augmented state design, the original non-stationary MDP becomes a stationary one. However, the alpha pool consists of complex mathematical expressions that are difficult to understand directly. This motivates us to use a large language model (LLM), together with a structured prompt, to encode the pool's semantic content. To address the single-objective reward issue, we define a multi-dimensional reward function that jointly characterizes the quality of alpha pool along four axes: predictive power (IC of the mega-alpha), temporal stability (consistency of mega-alpha over time), perturbation robustness (robustness of mega-alpha under market perturbations), and diversity (complementarity relative to existing pool). Rather than collapsing these rewards into a single scalar via manual weighting, we employ a Pareto-regularized learning method to generate alphas along the efficient frontier, producing principled trade-offs among all objectives.

In summary, our main contributions are three-fold:
\begin{enumerate}
    \item Conceptually, we reformulate the non-stationary alpha discovery MDP with a single-objective reward into a context-conditional stationary MDP with a multi-objective reward, better aligning the framework with practical challenges.
    
    \item Methodologically, we propose AlphaPareto, which uses an LLM to encode the evolving alpha pool as a part of state, thereby stationarizing the alpha discovery MDP, and introduces a Pareto-regularized learning procedure to obtain preference-conditioned optimal policies.

    \item Empirically, we conduct extensive experiments on large-scale real-world datasets from Chinese stock markets, demonstrating that AlphaPareto consistently outperforms state-of-the-art baselines. Ablation studies further confirm the complementary contributions of each proposed component.

\end{enumerate}

\section{Background and Related Work}

\paragraph{GP-based alpha discovery.}
Over the past decade, genetic programming (GP) has been the dominant approach to formulaic alpha discovery \citep{huataigp1,huataigp2,zhang2020autoalpha,alphaevolve}. It evolves expression trees using IC as the fitness measure. However, GP scales poorly as the number of operators and features grows. More importantly, it discovers alphas independently and does not directly optimize for synergy within the alpha pool, leading to poor practical performance. 

\paragraph{RL-based alpha discovery.}
AlphaGen \citep{alphagen} is the first to model formulaic alpha discovery as an MDP and to use MaskPPO to train an agent that generates a set of synergistic formulaic alphas. This marks an important step beyond GP, since it makes the generator aware of the performance of alpha pool via reward. \cite{alphaqcm} point out the non-stationarity in this MDP and propose AlphaQCM, a distributional RL method that uses quantiled conditional variance as an exploration bonus to reduce the impact of non-stationarity. AlphaForge \citep{alphaforge} improves the combination model by incorporating alpha's temporal performance into selection and dynamically adjusting weight for each alpha. Despite these advances, all existing RL-based methods share the aforementioned two limitations: (i) none analyze alpha pool to handle non-stationarity; (ii) all optimize a single scalar reward, and therefore ignore the inherently multi-dimensional nature of alpha pool quality.

\paragraph{LLM-based alpha discovery.}

The emergence of LLMs has opened a new direction for alpha discovery \citep{wang2025alphagpt, yuan2024alphagpt2, alphaagent, rdagent, luo2026alphabench}. Most of these methods use LLMs as \emph{factor generators} that directly produce formulaic alpha expressions. For example, AlphaAgent \citep{alphaagent} uses LLMs to propose market hypotheses and construct factors with regularization on originality and complexity. R\&D-Agent-Quant \citep{rdagent} separates this process into a Research stage, forming hypotheses from domain priors, and a Development stage, using code-generation agents for backtesting. However, \citet{luo2026alphabench} highlight that the robustness and search efficiency of LLM-based alpha discovery remain significant challenges, warranting further investigation. Rather than using the LLM as a direct alpha generator, AlphaPareto uses it as a semantic encoder of the alpha pool within an RL loop.

\section{Problem Setup}

We consider a market with $N$ stocks observed over $S$ trading days. Our ultimate goal is to discover an alpha pool $\calF = \{f^{1}, \ldots, f^{P}\}$ with best out-of-sample predictive power, containing $P$ formulaic alphas. Each alpha $f^{p}$ maps historical market data $\bm{H}_{s-1}$ to a cross-sectional signal $\balpha_{s}^{p} = f^{p}(\bm{H}_{s-1}) \in \RR^N$. These signals are then combined into a mega-alpha via a linear model:
\begin{align}\label{eq:mega_alpha}
\widehat{\balpha}_s(\calF) = \sum_{p=1}^{P} \widetilde{\balpha}_{s}^{p}\,\widehat{\beta}^{\,p} \in \RR^N,
\end{align}
where $\widetilde{\balpha}_{s}^p = \big(\balpha_{s}^{p} - \Mean(\balpha_{s}^{p})\big) / \Std(\balpha_{s}^{p})$ is the cross-sectionally standardized alpha signal, and $\widehat{\bbeta} = (\widehat{\beta}^{\,1}, \ldots, \widehat{\beta}^{\,P})^\top$ is estimated by minimizing $\sum_{s \in \training} \|\bm{y}_s - \widehat{\balpha}_s(\calF)\|^2$, with $\training$ being the training set and $\bm{y}_s \in \RR^N$ denoting future stock returns.

Next, following prior work, each formulaic alpha $f$ is represented in reverse Polish notation (RPN), so that its mathematical expression is encoded as a token sequence $z$, consisting of operators and features.\footnote{See Appendix \ref{sec:RPN} for more details about RPN and available tokens.} Since each $z$ is generated through a series of token selections, the discovery of $f$ can be formulated as a sequential decision-making problem. This process is modeled as an MDP $(\calX, \calA, \calP, \gamma, \calR)$, which is elaborated below.

Conventionally, each state $x_t \in \calX$ corresponds to a RPN token sequence $z_t$, with initial state $x_0 = z_0 \equiv \textit{BEG}$, where \textit{BEG} is a special token indicating the beginning of a formula. Each action $a_t \in \calA$ is a token selected by the RL agent. To ensure that $z_t$ is a valid RPN, only a subset of actions is allowed at each state, enforced through invalid action masking \citep{alphagen}. 

The transition kernel $\calP$ is deterministic: given $x_t$ and $a_t$, the next state $x_{t+1}$ is
\[
x_{t+1} = \calP(x_t, a_t) = z_{t+1} = [z_t, a_t].
\]
An episode terminates when $a_t = \textit{SEP}$ (a special token) or $|x_{t+1}| > L_{\max}$, where $L_{\max}$ is the maximum sequence length used to preserve interpretability. The discount factor is set to $\gamma = 1$, since each episode is finite in this MDP.

Let $f_t$ denote the parsed formulaic alpha corresponding to the next token sequence $z_{t+1}$. The reward $r_t$ measures the contribution of $f_t$ to the current alpha pool $\calF_t$. If $f_t$ is incomplete (or invalid), $r_t = 0$ (or $-1$); otherwise, (i) $f_t$ is temporarily added to $\calF_t$, forming an extended pool $\overline{\calF}_t$, (ii) a linear model is fitted on $\overline{\calF}_t$ and the alpha pool is updated to $\calF_{t+1}$ by keeping at most $P$ alphas based on their fitted contributions, and (iii) the reward is defined as its in-sample IC:
\begin{align}\label{eq:reward_classical}
\begin{split}
    r_t &= \IC(\{\widehat{\balpha}_{s}(\calF_{t+1}) : s \in \training\}) = \Mean(\Corr(\widehat{\balpha}_{s}(\calF_{t+1}), \bm{y}_s))
    \\
    &= \frac{1}{|\training|} \sum_{s \in \training} \frac{\sum_{i=1}^N \left[\widehat{\alpha}_{i,s}(\calF_{t+1}) - \bar{\alpha}_s(\calF_{t+1}) \right] (y_{i,s} - \bar{y}_s)} {\sqrt{\sum_{i=1}^N [\widehat{\alpha}_{i,s}(\calF_{t+1}) - \bar{\alpha}_s(\calF_{t+1})]^2 \sum_{i=1}^N (y_{i,s} - \bar{y}_s)^2}},
\end{split}
\end{align}
where $\widehat{\balpha}_{s}(\calF_{t+1})$ is the mega-alpha defined in \eqref{eq:mega_alpha}, $\bm{y}_s \in \RR^N$ is the vector of future returns, $\widehat{\alpha}_{i,s}$ and $y_{i,s}$ are the $i$-th elements of $\widehat{\balpha}_s(\calF_{t+1})$ and $\bm{y}_s$, respectively. Here, $\bar{\alpha}_s = \frac{1}{N}\sum_{i=1}^N \widehat{\alpha}_{i,s}$ and $\bar{y}_s = \frac{1}{N}\sum_{i=1}^N y_{i,s}$ are the cross-sectional averages of $\widehat{\balpha}_s(\calF_{t+1})$ and $\bm{y}_s$. Clearly, the IC measures the average cross-sectional correlation between mega-alphas and target returns over all $s \in \training$. Thus, $r_t$ reflects the in-sample predictive power of the updated alpha pool $\calF_{t+1}$ in terms of correlation.

\paragraph{Non-stationarity.}
Under this classical specification of MDP, the reward can be re-written as
\begin{align*}
    r_t = \calR(\calF_{t+1}) = \calR(f_t, \calF_t) = \calR(\phi(z_{t+1}), \calF_t) = \calR(\phi(x_{t+1}), \calF_t) = \calR(\phi(\calP(x_t, a_t)), \calF_t),
\end{align*}
where $\phi(\cdot)$ maps a token sequence to its formula. Hence, the reward function $\calR$ depends not only on the state-action pair $(x_t,a_t)$, but also on the current alpha pool $\calF_t$. Since $\calF_t$ evolves across episodes as new alphas are added and old ones are removed, the same state-action trajectory may receive different rewards at different stages of training. This makes the discovery process inherently non-stationary. A natural idea is to include $\calF_t$ in the state so that the agent can also condition on the current pool $\calF_t$. In practice, however, this is difficult because the alpha pool is a set of complex mathematical expressions whose shared structure and complementary information are hard to summarize, even for domain experts.

\paragraph{Single-objective reward design.}
The reward in \eqref{eq:reward_classical} depends only on IC, a single scalar measuring average in-sample predictive power. However, as aforementioned, the quality of an alpha pool is inherently multi-dimensional. A pool with high IC may still exhibit large temporal variation or be fragile under market regime shifts, leading to poor out-of-sample predictive performance. To improve generalization, other criteria, such as temporal stability, perturbation robustness, and diversity, should also be incorporated into the reward of this MDP.

\section{Methodology}\label{sec:method}

\begin{figure}[ht]
    \centering
    \includegraphics[width=\linewidth]{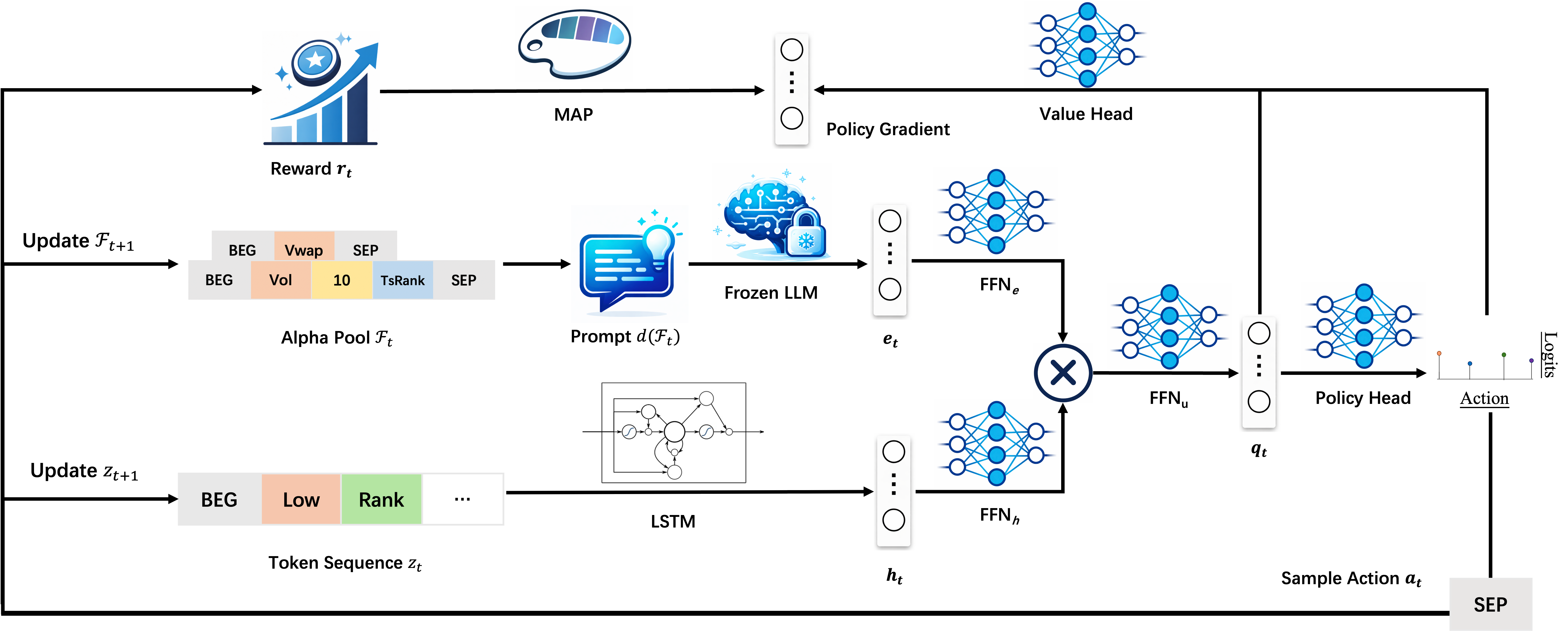}
    \caption{Overview of the AlphaPareto framework.}
    \label{fig:overview}
\end{figure}

In this section, we present the AlphaPareto method. Specifically, we first propose an alpha-pool semantic reasoning technique to augment the state representation and overcome non-stationarity. We then design a multi-objective vector-valued reward that simultaneously accounts for predictive power, temporal stability, perturbation robustness, and diversity, and introduce an optimization method to handle these competing objectives. An overview of AlphaPareto is shown in \cref{fig:overview}.

\subsection{Alpha Pool Semantic Reasoning}\label{sec:pool_reasoning}

A natural solution for addressing non-stationarity is to augment the state $x_t$ with the current alpha pool $\calF_t$ and redefine it as
\begin{align*}
    x_t = (z_t, \calF_t).
\end{align*}
However, directly encoding $\calF_t$ is difficult in practice, because it is a set of mathematical expressions whose synergy, redundancy, and missing aspects of market dynamics are hard to quantify explicitly. 

To overcome this difficulty, AlphaPareto uses a frozen LLM as a semantic encoder of $\calF_t$. Specifically, we construct a structured prompt $d(\calF_t)$ from the current alpha pool to summarize the mathematical expression and assigned weight of each alpha. This prompt is designed to help the LLM capture the pool's composition, concentration, and potential blind spots.\footnote{The full prompt template is provided in Appendix~\ref{sec:pool_prompt}.} Given $d(\calF_t)$, we extract the last-layer hidden representation of the LLM:
\begin{align}\label{eq:llm_emb}
    \bm{e}_t = \LLM(d(\calF_t)) \in \RR^{d_e},
\end{align}
where $\LLM(\cdot)$ and $d_e$ denote the forward mapping and hidden dimension of the frozen LLM, respectively. As a result, $\bm{e}_t$ is expected to provide a compact semantic summary of $\calF_t$, indicating which signals are already captured by the current pool and which complementary signals may still be missing. During RL training, the LLM remains frozen and no gradient is back-propagated through it.

Meanwhile, following \cite{alphagen}, the token sequence $z_t$ is encoded by an LSTM:
\begin{align}\label{LSTM_head}
    \bm{h}_t = \LSTM(z_t) \in \RR^{d_h},
\end{align}
where $d_h$ is the hidden dimension. We then project both $\bm{h}_t$ and $\bm{e}_t$ into a shared latent space through two feed-forward networks (FFNs), and combine them through element-wise product followed by another FFN:
\begin{align}\label{eq:ffns}
    \bm{q}_t = \FFN_u(\bm{u}_t)
    = \FFN_u \Big(\FFN_h(\bm{h}_t) \odot \FFN_e(\bm{e}_t)\Big)
    \in \RR^{d}.
\end{align}

Finally, the resulting representation $\bm{q}_t$ is fed into the policy and value heads of MaskPPO, as in \cite{alphagen}, to produce action probabilities and value estimates. As a result, the policy $\pi(a_t \mid x_t) = \pi(a_t \mid z_t, \calF_t)$ is conditioned on both the token sequence and the alpha pool, which helps resolve the non-stationarity caused by state--reward misalignment. See \cref{fig:overview} for an illustration.

\subsection{Pareto-Regularized MORL}\label{sec:morl}

\subsubsection{Multi-objective Reward}

To overcome the limitation of single-objective design, AlphaPareto replaces the scalar reward in \eqref{eq:reward_classical} with a vector-valued reward and adopts the Multi-Human-Value Alignment Palette (MAP; \citealp{wang2025map}) framework to optimize along the Pareto efficient frontier of competing objectives. 

Specifically, we measure the four-dimensional contribution of $f_t$ to the current alpha pool $\calF_t$: predictive power, temporal stability, perturbation robustness, and diversity. The predictive-power component is defined as
\begin{align*}
    \IC_t = \IC(\{\widehat{\balpha}_{s}(\calF_{t+1}) : s \in \training \}), 
\end{align*}
which is the same as the reward used in the classical MDP, defined in \eqref{eq:reward_classical}.

Then, temporal stability measures the consistency of the cross-sectional rankings produced by the mega-alpha over time. Its reward form is defined as
\begin{align*}
    \RRE_t
    &= \RRE(\{\widehat{\balpha}_{s}(\calF_{t+1}) : s \in \training\}) = \frac{1}{|\training|}\left( 1+\sum_{s=2}^{|\training|}
    \frac{1}{1 + D_{\mathrm{KL}}(\bm{p}_s \,\|\, \bm{p}_{s-1})} \right),
\end{align*}
where $\RRE(\cdot)$ denotes the Relative Rank Entropy (RRE), $D_{\mathrm{KL}}(\cdot \,\|\, \cdot)$ denotes the Kullback--Leibler (KL) divergence, and $\bm{p}_s = (p_{1,s}, \dots, p_{N,s})^\top$ is the cross-sectional ranking vector at time $s$, with 
\[
p_{i,s}
=
\frac{\Rank(\widehat{\alpha}_{i,s}(\calF_{t+1}))}
{\sum_{j=1}^N \Rank(\widehat{\alpha}_{j,s}(\calF_{t+1}))}.
\]
Clearly, a higher $\RRE_t$ indicates more stable stock rankings over time and therefore better temporal stability of the updated alpha pool $\calF_{t+1}$.

Next, perturbation robustness quantifies how robust the mega-alpha remains under perturbations. We define its reward component as
\begin{align*}
    \PFS_t & = \PFS(\{\widehat{\balpha}_{s}(\calF_{t+1}) : s \in \training\}) = \frac{1}{|\training|}\sum_{s \in \training} \rho \Big(\widehat{\balpha}_s(\calF_{t+1}),\; \widehat{\balpha}^\ast_{s} (\calF_{t+1})\Big),
\end{align*}
where $\PFS(\cdot)$ denotes the perturbation fidelity score (PFS), $\widehat{\balpha}_{s}^\ast(\calF_{t+1}) = (\bm{1}+\bm{\varepsilon}_s) \odot\widehat{\balpha}_s(\calF_{t+1})$ is the perturbed mega-alpha at time $s$, and $\rho(\cdot,\cdot)$ denotes the Spearman rank correlation. Here, the perturbation noise $\bm{\varepsilon}_s$ is independently sampled from a predefined distribution $\mathcal{E}$, where $\mathcal{E}$ is chosen to be either a Gaussian distribution or a Student-$t_3$ distribution with equal probability. Particularly, the Gaussian distribution is used to simulate random market microstructure fluctuations, while the heavy-tailed $t_3$ distribution captures structural shocks such as policy changes. These two distributions are rescaled so that their variances match the empirical cross-sectional variance of stock returns. Therefore, a higher $\PFS_t$ indicates that the alpha pool is more robust to disturbances and thus more likely to generalize well under changing market conditions.\footnote{\citet{ding2025alphaeval} define PFS by perturbing the stock features $\bm{H}_{s-1}$ and then taking the minimum over the two perturbation types, which makes the resulting measure more fragile and indirect.}

Lastly, diversity captures whether the alpha pool contains varied and complementary signals. We quantify the reward component of diversity through the diversity entropy (DH):
\begin{align*}
    \mathrm{DH}_t = \mathrm{DH} (\{\widehat{\balpha}_{s}(\calF_{t+1}) : s \in \training\}) = -\frac{\sum_{i=1}^{|\calF_{t+1}|} \tilde{p}_i \log \tilde{p}_i}{\log |\calF_{t+1}|},
\end{align*}
where $\tilde{p}_i = \lambda_i / \sum_{j = 1}^{|\calF_{t+1}|} \lambda_j$, and $\lambda_1, \ldots, \lambda_{|\calF_{t+1}|}$ are the eigenvalues of the covariance matrix $\bm{C} \in \RR^{|\calF_{t+1}| \times |\calF_{t+1}|}$ computed from the individual alpha signals $\{\widehat{\balpha}_{s}^p: p = 1, \dots, |\calF_{t+1}|, s \in \training\}$. Concretely, a higher $\mathrm{DH}_t$ indicates more evenly distributed variance across orthogonal directions, and thus $\calF_{t+1}$ contains more complementary information and less redundancy among its constituents.

Combining all these components, we obtain the vector-valued reward for multi-objective learning:
\begin{align}\label{vector_reward}
    \bm{r}_t = (\IC_t, \RRE_t, \PFS_t, \mathrm{DH}_t)^\top.
\end{align}
Conventionally, for incomplete (or invalid) formulas, we set $\bm{r}_t=\bm{0}$ (or $-\bm{1}$) and evaluate the above quantities only when a valid formula is terminated.

\subsubsection{MAP-based Optimization}

Realistically, the four objectives in \eqref{vector_reward} may conflict. For example, improving predictive power may reduce diversity, since focusing on a narrow class of alphas can work better in a given market condition. It is therefore essential to consider the Pareto frontier of this multi-objective optimization. The MAP framework shows that a linear combination of individual reward functions is sufficient to reach the Pareto frontier under mild assumptions. Motivated by this finding, we collapse the competing rewards into a single scalar with data-driven weights, under some realistic constraints.

Specifically, let $\bc = (c_{\IC}, c_{\RRE}, c_{\PFS}, c_{\mathrm{DH}})^\top$ denote a vector of preferred levels, where each entry specifies the desired minimum level for one objective. We refer to $\bc$ as the \emph{value palette}. By Theorem 2 of \citet{wang2025map}, the Pareto-regularized optimization under vector-valued return $\bm{r}_t$ is equivalent to optimizing the linear-scalarized reward:
\begin{align}\label{eq:scalarized_reward}
    r^\ast_t = \blam(\bc)^\top \bm{r}_t,
\end{align}
where the weight vector $\blam(\bc) \in \RR^4$ is the unique maximizer of the dual objective: 
\begin{align}\label{eq:map_dual}
    \argmax_{\blam \geq \bm{0}} \, g(\bm{\lambda}; \bm{c}) = \argmax_{\blam \geq \bm{0}}
    -\log \EE_{\pi}\!\left[\exp\!\big(\bm{\lambda}^\top \bm{r}_t \big)\right]
    + \bm{\lambda}^\top \bm{c}.
\end{align}
Clearly, $\blam(\bc)$ depends on both the multi-objective reward $\bm{r}_t$ under policy $\pi$ and the chosen value palette $\bc$. In this sense, $\bc$ acts as a regularizer that encodes the user's preferences over the multiple objectives, while $\blam(\bc)$ adaptively balances them according to the current dynamics. This formulation requires only a minimum target level for each objective in $\bc$, rather than fully subjective scalarization weights, and thus provides a principled way to trade off these competing objectives.

Moreover, MAP is originally developed for an offline setting with fixed transitions. We slightly extend it to our online alpha-discovery setting by recomputing the palette-dependent multiplier $\blam(\bc)$ in the online manner. Particularly, given $T$ sampled transitions $\{(x_t, a_t, \bm{r}_t, x_{t+1})\}$ from the replay buffer, we solve the empirical dual
\begin{align}\label{eq:empirical_dual}
    \widehat{\bm{\lambda}}(\bm{c}) =
    \argmax_{\bm{\lambda} \geq \bm{0}}
    \left\{
    -\log \frac{1}{T}\sum_{t=1}^{T}\exp\!\big(\bm{\lambda}^\top \bm{r}_t\big)
    + \bm{\lambda}^\top \bm{c}
    \right\}.
\end{align}
We then use the online scalarized reward $r_t^\ast(\bm{c}) = \widehat{\blam}(\bm{c})^\top \bm{r}_t$ inside MaskPPO and update the policy and value networks in the standard way. 

Overall, the complete AlphaPareto algorithm is summarized in \cref{alg:alphapareto}.

\section{Experiments}\label{sec:experiments}

Conventionally, we conduct experiments on Chinese A-share stock market data, with the prediction target defined as the 20-day future stock return. To examine how market complexity affects alpha discovery, we consider three stock universes of increasing difficulty: (1) the largest 300 stocks (CSI300), (2) the largest 800 stocks (CSI800), and (3) the full market universe (Market). As the stock universe expands, discovering a synergistic alpha pool becomes more challenging because the underlying market dynamics become more complex. Each dataset is split chronologically into a training period (2013/07/01--2023/06/30), a validation period (2023/07/01--2024/06/30), and a test period (2024/07/01--2025/06/30).

We compare AlphaPareto with the following four categories of baselines:
\begin{enumerate}
    \item[(1)] \textbf{Alpha158} and \textbf{MLP} (handcrafted formulaic and end-to-end non-formulaic baselines): \textbf{Alpha158} fixes the alpha pool to the 158 predefined formulaic alphas in Qlib \citep{qlib} and combines them with a linear model to form a mega-alpha, whereas \textbf{MLP} replaces the linear model with a multi-layer perceptron network.
    \item[(2)] \textbf{GP} (GP-based formulaic alpha discovery): \textbf{GP} uses genetic programming to generate candidate alphas and builds the mega-alpha from the top-$P$ alphas.
    \item[(3)] \textbf{AlphaAgent} and \textbf{R\&D-Agent-Quant} (LLM-based formulaic alpha discovery): These methods use LLMs as alpha generators to propose candidate factors or formulas, from which the final alpha pool is constructed.
    \item[(4)] \textbf{AlphaGen}, \textbf{AlphaForge} and \textbf{AlphaQCM} (RL-based formulaic alpha discovery): These methods use RL to search for a synergistic alpha pool and then fit a linear mega-alpha based on the discovered formulas.
\end{enumerate}

Each stochastic experiment is repeated with $5$ random seeds, and the alpha-pool size $P$ is tuned on the validation set. Since LLM-based alpha discovery methods heavily rely on the underlying LLM's capability, AlphaAgent and R\&D-Agent-Quant use the largest DeepSeek-V3.2 (671B) model as the alpha generator. In contrast, AlphaPareto uses the LLM as a semantic encoder to obtain the hidden state, and we choose the Qwen-Embedding-4B model for this purpose. More details on the baseline methods are provided in Appendix \ref{sec:baseline}, and hyperparameters for AlphaPareto are listed in Appendix \ref{sec:hyper}. Additional experiments on the impact of MAP, random-noise control, and cross-market generalization are provided in the Appendices \ref{sec:morl_comparison}--\ref{sec:sp500}. The source code for reproducing the experiments is available at \url{https://github.com/BiQiBaoWinner/AlphaPareto}.

\subsection{Main Comparison}

We first assess how different alpha discovery methods affect out-of-sample predictive performance, since predictive power is the most important criterion in practice. Table \ref{tab:ic_method} reports the out-of-sample ICs on the CSI300, CSI800, and Market datasets. From this table, AlphaPareto achieves the highest IC on all three datasets, reaching $3.92\%$ on CSI300, $5.70\%$ on CSI800, and $10.10\%$ on Market, outperforming all competing methods: $3.69\%$, $5.07\%$ and $9.16\%$ earned by the second best method. The paired $t$-tests reported in Appendix \ref{sec:paired_tests} provide further statistical support for the advantages of AlphaPareto.

The gain of AlphaPareto over the strongest RL-based baseline is moderate on CSI300 and CSI800 ($0.23\%$ and $0.63\%$), but becomes much larger on the Market dataset ($0.94\%$). This pattern suggests that conditioning on the evolving alpha pool and optimizing multiple objectives may be especially useful in complex markets, where discovering synergistic alphas is more difficult.

Meanwhile, the LLM-based baselines do not perform well in this setting. A possible reason is that out-of-sample market conditions can differ substantially from the in-sample period, making it difficult for direct LLM generation to produce stable and effective alphas. In contrast, AlphaPareto uses the LLM only to encode the alpha pool and support decision making within the RL loop.

\begin{table}[!h]
  \centering
  \caption{Out-of-sample ICs of different methods across different datasets.}
  \setlength{\tabcolsep}{2.7mm}
    \begin{tabular}{ccccccccc}
    \toprule
          & \multicolumn{2}{c}{CSI300} &       & \multicolumn{2}{c}{CSI800} &       & \multicolumn{2}{c}{Market} \\
\cmidrule{2-3}\cmidrule{5-6}\cmidrule{8-9}    Method & Mean  & Std   &       & Mean  & Std   &       & Mean  & Std \\
\cmidrule{1-3}\cmidrule{5-6}\cmidrule{8-9}    Alpha158 & 2.99\% & \textcolor[rgb]{ .129,  .361,  .596}{-} &       & 4.77\% & \textcolor[rgb]{ .129,  .361,  .596}{-} &       & 4.04\% & \textcolor[rgb]{ .129,  .361,  .596}{-} \\
    MLP   & 1.51\% & \textcolor[rgb]{ .129,  .361,  .596}{(2.33\%)} &       & 4.63\% & \textcolor[rgb]{ .129,  .361,  .596}{(0.90\%)} &       & 7.00\% & \textcolor[rgb]{ .129,  .361,  .596}{(0.18\%)} \\
    
    GP    & 1.09\% & \textcolor[rgb]{ .129,  .361,  .596}{(0.82\%)} &       & 2.74\% & \textcolor[rgb]{ .129,  .361,  .596}{(1.40\%)} &       & 7.52\% & \textcolor[rgb]{ .129,  .361,  .596}{(1.53\%)} \\
    AlphaAgent & 0.51\% & \textcolor[rgb]{ .129,  .361,  .596}{(0.34\%)} &       & 0.48\% & \textcolor[rgb]{ .129,  .361,  .596}{(0.46\%)} &       & 1.10\% & \textcolor[rgb]{ .129,  .361,  .596}{(0.59\%)} \\
    R\&D-Agent-Quant & 2.39\% & \textcolor[rgb]{ .129,  .361,  .596}{(0.96\%)} &       & 2.45\% & \textcolor[rgb]{ .129,  .361,  .596}{(0.73\%)} &       & 1.48\% & \textcolor[rgb]{ .129,  .361,  .596}{(0.91\%)} \\
    AlphaGen & 3.69\% & \textcolor[rgb]{ .129,  .361,  .596}{(0.66\%)} &       & 5.07\% & \textcolor[rgb]{ .129,  .361,  .596}{(0.63\%)} &       & 8.44\% & \textcolor[rgb]{ .129,  .361,  .596}{(1.09\%)} \\
    AlphaQCM & 0.50\% & \textcolor[rgb]{ .129,  .361,  .596}{(0.95\%)} &       & 4.79\% & \textcolor[rgb]{ .129,  .361,  .596}{(0.34\%)} &       & 9.16\% & \textcolor[rgb]{ .129,  .361,  .596}{(4.22\%)} \\
    AlphaForge & 2.03\% & \textcolor[rgb]{ .129,  .361,  .596}{(0.65\%)} &       & 4.17\% & \textcolor[rgb]{ .129,  .361,  .596}{(1.96\%)} &       & 5.37\% & \textcolor[rgb]{ .129,  .361,  .596}{(1.81\%)} \\
    \textbf{AlphaPareto} & \pmb{3.92\%} & \textcolor[rgb]{ .129,  .361,  .596}{\pmb{(0.46\%)}} &       & \pmb{5.70\%} & \textcolor[rgb]{ .129,  .361,  .596}{\pmb{(0.87\%)}} &       & \pmb{10.10\%} & \textcolor[rgb]{ .129,  .361,  .596}{\pmb{(1.01\%)}} \\
    
    \bottomrule
    \end{tabular}%
  \label{tab:ic_method}%
\end{table}%

\begin{figure}[!h]
    \centering
    \includegraphics[width=0.9\linewidth]{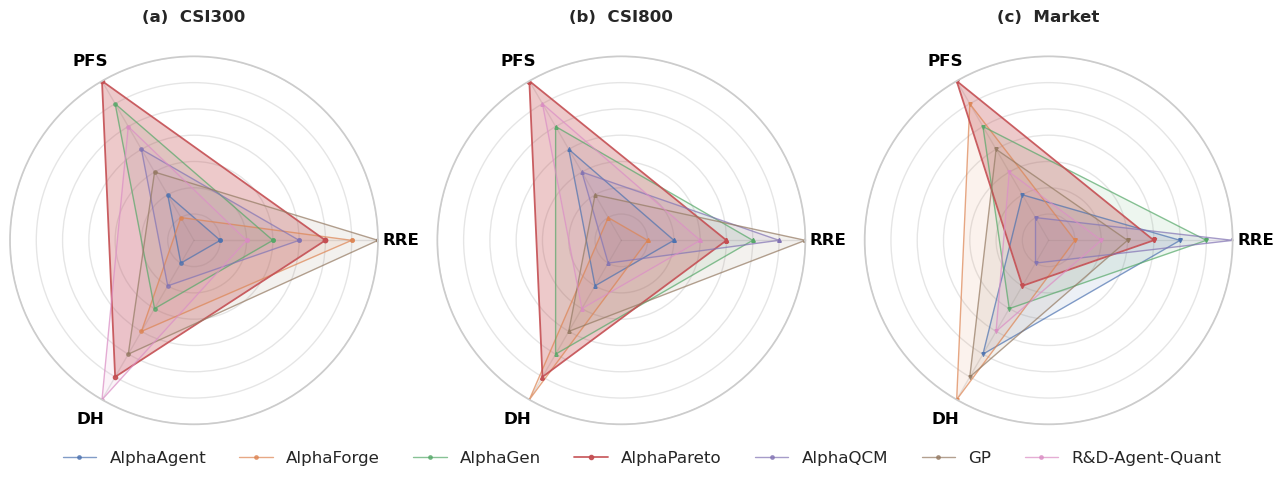}
    \caption{Radar chart comparison in terms of temporal stability, perturbation robustness, and diversity.}
    \label{fig:other_metrics}
\end{figure}

Beyond predictive performance, \cref{fig:other_metrics} compares the out-of-sample results in terms of temporal stability, perturbation robustness, and diversity. The figure shows that, together with having the highest out-of-sample predictive power, AlphaPareto also performs strongly across these three dimensions. In particular, on CSI300 and CSI800, it achieves the highest perturbation robustness (PFS) and the second-highest diversity (DH), while maintaining competitive temporal stability (ranking third or fourth in RRE). On the Market dataset, AlphaPareto still attains the highest PFS, though its DH and RRE are relatively moderate. Overall, these results further support the design of AlphaPareto: by employing multi-objective optimization, it discovers alpha pools that are not only more predictive, but also more stable, robust, and diverse. 

We further evaluate the out-of-sample economic performance of AlphaPareto through portfolio backtesting.\footnote{Details of the trading strategy and additional portfolio results are provided in Appendix \ref{sec:portfolio}.}\cref{tab:monthly_portfolio} summarizes the primary portfolio results. From this table, AlphaPareto achieves the highest annualized return, information ratio, and Sharpe ratio among all evaluated methods, with an annualized return of $58.92\%$, an information ratio of $2.62$, and a Sharpe ratio of $2.59$. Its maximum drawdown is $-18.12\%$, which is slightly worse than those of GP and AlphaGen but better than remaining methods. Overall, AlphaPareto provides the strongest portfolio-level performance, suggesting that its improved alpha discovery translates into more effective portfolio construction.

\begin{table}[!h]
  \centering
  \caption{Out-of-sample monthly portfolio performance. AV denotes the annualized return, MDD denotes the maximum drawdown, IR denotes the information ratio, and SR denotes the Sharpe ratio.}
  \label{tab:monthly_portfolio}
  \setlength{\tabcolsep}{6.7mm}
  \begin{tabular}{ccccc}
    \toprule
    Model & AV ($\uparrow$) & MDD ($\uparrow$) & IR ($\uparrow$) & SR ($\uparrow$) \\
    \midrule
    Alpha158 & 33.17\% & -18.71\% & 1.08  & 1.34  \\
    MLP & 47.09\% & -25.21\% & 1.57  & 1.58  \\
    GP & 24.49\% & \pmb{-13.29\%} & 0.77  & 1.18  \\
    AlphaGen & 40.03\% & -15.91\% & 1.65  & 1.61  \\
    AlphaAgent & 35.07\% & -29.94\% & 0.71  & 0.89  \\
    R\&D-Agent-Quant & 37.30\% & -30.18\% & 0.76  & 0.96  \\
    AlphaQCM & 19.45\% & -19.74\% & 0.20  & 1.21  \\
    AlphaPareto & \pmb{58.92\%} & -18.12\% & \pmb{2.62}  & \pmb{2.59}  \\
    \bottomrule
  \end{tabular}%
\end{table}

\subsection{Scaling Performance}

The choice of LLM within AlphaPareto tends to affect its performance. To examine this effect, we vary the parameter scale of the frozen LLM while keeping the rest of the framework unchanged. We use the DeepSeek-R1 model family to provide LLMs of different parameter scales. Table \ref{tab:ic_scaling} reports the out-of-sample ICs of AlphaPareto with different LLM scales. The results show that AlphaPareto presents strong performance over a wide range of scales, but the pattern is clearly non-monotonic. Notably, the Qwen-Embedding-4B model achieves the best ICs on all datasets.

This finding suggests that the role of the LLM in AlphaPareto differs from that in direct generation-based methods. Here, the LLM serves only as a frozen semantic encoder of the alpha pool rather than a generator of new alphas. In this setting, a small or medium-sized model may already provide sufficient semantic information, whereas excessively large models may introduce unnecessary domain knowledge or noise into the representation of alpha pool. From a practical perspective, these scaling results are encouraging, as they make AlphaPareto more cost-effective for real-world deployment.

\begin{table}[!h]
  \centering
  \caption{Out-of-sample ICs of AlphaPareto across different LLM sizes.}
  \setlength{\tabcolsep}{2.4mm}
    \begin{tabular}{ccccccccc}
    \toprule
          & \multicolumn{2}{c}{CSI300} &       & \multicolumn{2}{c}{CSI800} &       & \multicolumn{2}{c}{Market} \\
\cmidrule{2-3}\cmidrule{5-6}\cmidrule{8-9}    Model & Mean  & Std   &       & Mean  & Std   &       & Mean  & Std \\
\cmidrule{1-3}\cmidrule{5-6}\cmidrule{8-9}    
    DeepSeek-1.5B  & 2.65\% & \textcolor[rgb]{ .129,  .361,  .596}{(0.99\%)} &       & 5.39\% & \textcolor[rgb]{ .129,  .361,  .596}{(0.44\%)} &       & 9.33\% & \textcolor[rgb]{ .129,  .361,  .596}{(0.92\%)} \\
    \pmb{Qwen-Embedding-4B}    & \pmb{3.92\%} & \textcolor[rgb]{ .129,  .361,  .596}{\pmb{(0.46\%)}} &       & \textbf{5.70\%} & \textcolor[rgb]{ .129,  .361,  .596}{\pmb{(0.87\%)}} &       & \pmb{10.10\%} & \textcolor[rgb]{ .129,  .361,  .596}{\pmb{(1.01\%)}} \\
    DeepSeek-7B    & 3.38\% & \textcolor[rgb]{ .129,  .361,  .596}{(0.56\%)} &       & 5.50\% & \textcolor[rgb]{ .129,  .361,  .596}{(1.23\%)} &       & 9.50\% & \textcolor[rgb]{ .129,  .361,  .596}{(0.56\%)} \\
    DeepSeek-14B   & 3.71\% & \textcolor[rgb]{ .129,  .361,  .596}{(1.73\%)} &       & 5.59\% & \textcolor[rgb]{ .129,  .361,  .596}{(0.46\%)} &       & 9.74\% & \textcolor[rgb]{ .129,  .361,  .596}{(0.64\%)} \\
    DeepSeek-32B   & 3.26\% & \textcolor[rgb]{ .129,  .361,  .596}{(1.43\%)} &       & 4.76\% & \textcolor[rgb]{ .129,  .361,  .596}{(0.52\%)} &       & 9.88\% & \textcolor[rgb]{ .129,  .361,  .596}{(1.13\%)} \\
    \bottomrule
    \end{tabular}%
  \label{tab:ic_scaling}%
\end{table}%

\subsection{Ablation Study}

To examine the contribution of the two key components in AlphaPareto, we conduct an ablation study with four variants: the baseline without either component (i.e., AlphaGen), the model with only LLM-based alpha pool semantic reasoning, the model with only MAP-based MORL, and the full AlphaPareto method with both components. Table \ref{tab:ablation_study} reports the out-of-sample IC performance in this ablation study.

The results show that both components are useful, but they contribute in different ways. The LLM-based alpha pool semantic reasoning mainly improves performance on the Market dataset, where the underlying financial system is more complex. In contrast, the MAP-based MORL component improves performance on all three datasets, highlighting the value of improving the optimization target beyond predictive power alone. When the two components are combined, the resulting AlphaPareto method achieves the best performance on CSI800 and Market, while remaining very close to the best mean result on CSI300 and exhibiting substantially smaller variation (only a $0.07\%$ loss in mean, but a $0.81\%$ reduction in standard deviation). This slight decline in IC on CSI300 may arise from additional domain noise introduced by the LLM. The general market knowledge from the LLM can be useful for a broad stock universe, but may be less informative for the large, relatively efficiently priced stocks in CSI300. Consistent with this hypothesis, using the LLM component alone leads to a more pronounced performance decline on CSI300.

\begin{table}[!h]
  \centering
  \caption{Ablation study of AlphaPareto, where ``LLM'' denotes the LLM-based alpha pool semantic reasoning, and ``MORL'' denotes the MAP-based multi-objective reinforcement learning.}
  \setlength{\tabcolsep}{2.5mm}
    \begin{tabular}{cccccccccc}
    \toprule
    \multicolumn{2}{c}{Included Components} & \multicolumn{2}{c}{CSI300} &       & \multicolumn{2}{c}{CSI800} &       & \multicolumn{2}{c}{Market} \\
\cmidrule{1-4}\cmidrule{6-7}\cmidrule{9-10}    LLM   & MORL  & Mean  & Std   &       & Mean  & Std   &       & Mean  & Std \\
\cmidrule{1-4}\cmidrule{6-7}\cmidrule{9-10}          &       & 3.69\% & \textcolor[rgb]{ .129,  .361,  .596}{(0.66\%)} &       & 5.07\% & \textcolor[rgb]{ .129,  .361,  .596}{(0.63\%)} &       & 8.44\% & \textcolor[rgb]{ .129,  .361,  .596}{(1.09\%)} \\
    $\checkmark$ &       & 2.37\% & \textcolor[rgb]{ .129,  .361,  .596}{(0.68\%)} &       & 5.04\% & \textcolor[rgb]{ .129,  .361,  .596}{(0.42\%)} &       & 9.68\% & \textcolor[rgb]{ .129,  .361,  .596}{(0.83\%)} \\
          & $\checkmark$ & \pmb{3.99\%} & \textcolor[rgb]{ .129,  .361,  .596}{\pmb{(1.27\%)}} &       & 5.54\% & \textcolor[rgb]{ .129,  .361,  .596}{(1.03\%)} &       & 9.63\% & \textcolor[rgb]{ .129,  .361,  .596}{(0.80\%)} \\
    $\checkmark$ & $\checkmark$ & 3.92\% & \textcolor[rgb]{ .129,  .361,  .596}{(0.46\%)} &       & \pmb{5.70\%} & \textcolor[rgb]{ .129,  .361,  .596}{\pmb{(0.87\%)}} &       & \pmb{10.10\%} & \textcolor[rgb]{ .129,  .361,  .596}{\pmb{(1.01\%)}} \\
    \bottomrule
    \end{tabular}%
  \label{tab:ablation_study}%
\end{table}%

\section{Conclusion}\label{sec:conclusion}

We propose AlphaPareto, an RL framework for formulaic alpha discovery that addresses two key limitations of existing RL-based methods: non-stationarity and single-objective reward design. The method conditions the policy on an LLM-based representation of the evolving alpha pool and optimizes predictive power, temporal stability, perturbation robustness, and diversity simultaneously through Pareto-regularized multi-objective learning. Experiments on Chinese and U.S. data show that AlphaPareto outperforms existing baselines across multiple dimensions, while the scaling and ablation studies further indicate that strong performance does not require the largest available LLMs and that both proposed components play complementary roles.

Several limitations are worth noting. First, performance may depend on the prompt template, which has not been systematically studied. Second, we use a linear mega-alpha combiner, which may exclude useful nonlinear structures that could further improve performance.

\bibliographystyle{icml2026}
\bibliography{reference}


\appendix

\section{Reverse Polish Notation of Formulaic Alpha}\label{sec:RPN}

\begin{figure}[!h]
    \centering
    \includegraphics[width=\textwidth]{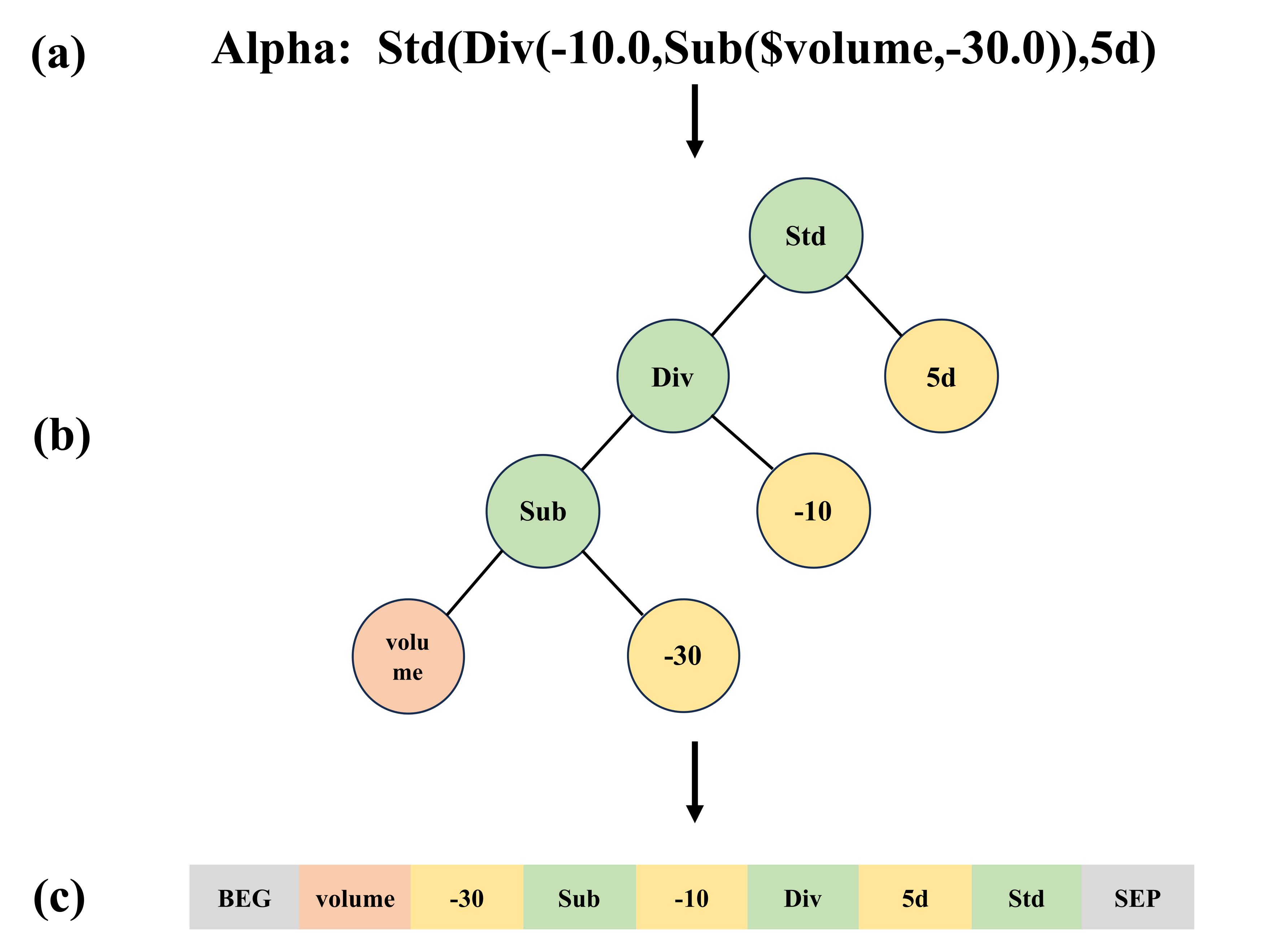}
    \caption{(a) A formulaic alpha discovered by our AlphaPareto method. (b) Its expression tree. (c) Its RPN representation.}
    \label{fig:RPN}
\end{figure}

In this section, we introduce the reverse Polish notation (RPN) used to represent formulaic alphas. \cref{fig:RPN} provides an illustrative example. In the figure, a formulaic alpha is encoded as a token sequence, where each feature and operator is represented as a token. Moreover, special tokens \textit{BEG} and \textit{SEP} denote the beginning and end of the expression, respectively.

\cref{tab:operators} lists all features and operators used in this paper. Following \cite{alpha101} and \cite{huataigp1,huataigp2}, these operators can be broadly divided into time-series operators and cross-sectional operators. Time-series operators require observations over multiple days, while cross-sectional operators act on single-day data. By combining these two types of operators, formulaic alphas can capture rich nonlinear patterns while remaining interpretable.

\begin{table}[!ht]
	\centering
	\caption{Description of available features and operators.}
    \resizebox{\columnwidth}{!}{
    \begin{tabular}{l|l}
        \toprule
        \textbf{Tokens} & \textbf{Description} \\
        \midrule
        \multicolumn{2}{c}{Features}\\
        \midrule
        Open/High/Low/Close/Vwap/Volume & Opening/high/low/closing/vwap price or volume of stock $i$ at time $s$.\\
        Constant & Number from $\{-30, -10, -5, -2, -1, -0.5, -0.01, 0.01, 0.5, 1, 2, 5, 10, 30\}$.\\ 
        Time delta & Integer from $\{10, 20, 30, 40, 50\}$, which is used in the time-series operators.\\
        \midrule
        \multicolumn{2}{c}{Time-series Operators}\\
        \midrule
        $\textit{Ref}\,(u_{i,s}, d)$ & Return the value of $u_{i, s - d}$, where $d$ is the time delta and $u_{i,s}$ is a feature of \\
        & stock $i$ at time $s$.\\
        $\textit{TsRank}\,(u_{i,s}, d)$ & Return the rank of $u_{i,s}$ among $\{u_{i,s}, \dots, u_{i, s-d}\}$\\
        $\textit{Mean}/\textit{Med}/\textit{Sum}/\textit{Std}/\textit{Var}\,(u_{i,s}, d)$ & Return the mean/median/sum/standard deviation/variance of $\{u_{i,s}, \dots, u_{i, s-d}\}$.\\
        $\textit{Max}/\textit{Min}\,(u_{i,s}, d)$ & Return the maximum/minimum value of $\{u_{i,s}, \dots, u_{i, s-d}\}$.\\
        $\textit{WMA}/\textit{EMA}\,(u_{i,s}, d)$ & Return the weighted/exponentially weighted moving average of $\{u_{i,s}, \dots,$ \\
        & $u_{i, s-d}\}.$\\
        $\textit{Cov}/\textit{Corr}\,(u_{i,s}, z_{i,s}, d)$ & Return the covariance/correlation between $u_{i,s}$ and $z_{i,s}$ based on samples \\
        & $\{(u_{i,s},z_{i,s}) \dots, (u_{i, s-d}, z_{i, s-d})\}$, where $u_{i,s}$ and $z_{i,s}$ are features of stock\\
        & $i$ at time $s$.\\
        \midrule
        \multicolumn{2}{c}{Cross-sectional Operators}\\
        \midrule
        $\textit{Sign}\,(u_{i,s})$ & Return 1 if the value of $u_{i,s}$ is positive, otherwise return 0.\\
        $\textit{Abs}\,(u_{i,s})$& Return the absolute value of $u_{i,s}$.\\
        $\textit{Log}\,(u_{i,s})$ & Return the logarithmic value of $u_{i,s}$.\\
        $\textit{Rank}\,(u_{i,s})$ & Return the rank of $u_{i,s}$ among $\{u_{1,s}, \dots, u_{N,s}\}$. \\
        $\textit{Add}/\textit{Sub}/\textit{Mul}/\textit{Div}\, (u_{i,s}, z_{i, s})$ & Return the result of adding/subtracting/multiplying/dividing $u_{i,s}$ and $z_{i, s}$. \\
        $\textit{Greater}/\textit{Less}\,(u_{i,s}, z_{i, s})$ & Return the greater/less value of $u_{i,s}$ and $z_{i, s}$.\\
        \bottomrule
    \end{tabular}\label{tab:operators}
    }
\end{table}

\section{Prompt Template}\label{sec:pool_prompt}

To construct the semantic representation of the current alpha pool, we feed a structured prompt to a frozen LLM. The prompt is designed to summarize the current pool in terms of factor expressions, weights, information coverage, and potential blind spots. The placeholder ``\$Current\_Alpha\_Pool.json\$'' denotes a JSON file containing the expressions and fitted weights of the current alpha pool $\calF_t$.

\begin{footnotesize}
\begin{Verbatim}[breaklines=true,breakanywhere=true]
[Role Definition] 
You are the Chief Investment Officer of a quantitative hedge fund managing over $5 billion in assets, with 20+ years of factor research experience. You excel at decoding market microstructure signals from mathematical expressions, precisely identifying the economic logic behind factors (e.g., liquidity premiums, behavioral bias arbitrage, volatility surface arbitrage), and evaluating their risk-return contributions in portfolios.

[Core Task]
Deeply internalize the mathematical structure and portfolio characteristics of the following factor pool. Your analysis must strictly derive from the JSON data—no external assumptions permitted. Focus on:
1) Factor Style Deconstruction: Map each expression to professional factor taxonomies (e.g., Barra style factors/alternative data factors/tail risk factors)
2) Information Topology Analysis: Detect data source blind spots (e.g., over-reliance on raw prices? missing volume-price interactions? ignoring volatility surfaces?)
3) Risk Heatmap: Identify vulnerabilities exposed by weight configuration (e.g., negative weight concentration, extreme value sensitivity, expression homogenization risk)
4) Evolution Path Design: Conceive next-generation factors that penetrate current information blind spots (specify mathematical operations and data types)

[Professional Analysis Framework]
Apply the following quantitative expertise for deep decoding:
- Price Transformation Layer: Identify market state assumptions implied by operations like Log/EMA/Greater/Less (e.g., Log($price) assumes log-normal distribution)
- Time Scale Layer: Parse window parameters (40d/20d/10d) to determine market cycle sensitivity (short-term sentiment vs. long-term trends)
- Nonlinearity Layer: Detect strategy logic formed by combinations of Add/Sub/Mul/Div with conditional functions (Greater/Less) (e.g., volatility breakout strategies)
- Weight Diagnostics: Calculate weight entropy to assess diversification; flag factors with |weight|>0.2 as risk concentration points
- Data Source Audit: Count frequencies of $high/$low/$close/$open/$vwap/$volume appearances to evaluate price dimension coverage completeness

[Critical Example Anchoring]
Below is the typical structure of the factor pool you will analyze (note expression complexity and weight distribution characteristics):
exprs = [
  'Sub(Mul(EMA(Log(Sum(Add(0.5,Div($high,0.01)),40d)),40d),-10.0),-0.01)',
  'Add(Div(Less(-30.0,$close),Add(30.0,Greater($low,-30.0))),Greater(Log($open),10.0))',
  'Sub(Less(Greater(Add(5.0,Add($close,-30.0)),-30.0),-10.0),-1.0)',
  'Div(2.0,$vwap)',
  'Sub(Div(5.0,$high),5.0)',
  'Add(Div(-10.0,$low),-5.0)',
  'Add(Sub(Abs(Log(Sub(Div(Greater(-5.0,Log($close)),0.01),10.0))),-0.5),-30.0)',
  'Greater(Div(EMA(Div(Greater(Log(Div(0.01,$high)),-30.0),-10.0),20d),$low),-10.0)',
  'Log(Sub(Greater(-30.0,$vwap),Div(Div($volume,$open),-30.0)))',
  'Ref(Log(Greater(Mul(-5.0,Mean(Mul($open,-0.5),10d)),Less(-30.0,$open))),20d)'
],
weights = [
  -0.0732,
  -0.2274,
  -0.0497,
   0.0456,
   0.2256,
  -0.0793,
  -0.1027,
  -0.4176,
  -0.1442,
   0.0939
],
>>> This example reveals critical risk: Factor #8 holds 41.76% absolute weight concentration with triple-nested nonlinear operations, highly vulnerable during market regime shifts

[Final Output Directive]
Now, channel your professional cognition entirely onto the real-world factor pool described in the JSON below.
OMIT all intermediate analysis steps, background context, and pleasantries.
Directly based on your analysis of current portfolio blind spots and risks, output 3 construction style recommendations for the "Next Factor," ranked by priority (High to Low).

Output format must strictly adhere to the following:
Allowed Style List: [Trend Momentum], [Mean Reversion], [Volatility Structure], [Price-Volume Interaction], [Liquidity Premium], [Market Sentiment], [Tail Risk], [Correlation Structure], [Higher-Moment Features], [Term Structure], [Asymmetry]

1. [Selected Style Name]: <Specific mathematical construction proposal, including recommended operators and data sources> —— <Expected contribution to filling current blind spots or reducing correlation>
2. [Selected Style Name]: ...
3. [Selected Style Name]: ...

[Cognitive Activation]
Begin decoding and output recommendations directly:

JSON:
$Current_Alpha_Pool.json$

\end{Verbatim}
\end{footnotesize}

\section{Algorithm}

Algorithm \ref{alg:alphapareto} presents the pseduo code for the AlphaPareto method.

\begin{algorithm}[!h]
\caption{AlphaPareto Training Pipeline}
\label{alg:alphapareto}
\begin{algorithmic}[1]
\REQUIRE Training data $\{(\bm{H}_{s-1}, \bm{y}_s)\}_{s \in \training}$, frozen LLM $\LLM(\cdot)$, max pool size $P$, max formula length $L_{\max}$, value palette $\bm{c}$, update period $\overline{n}$, and max steps $n^\ast$
\STATE Initialize alpha pool $\calF \leftarrow \emptyset$, replay buffer $\mathcal{B} \leftarrow \emptyset$, $n \leftarrow 0$, policy $\pi_\theta$, and value network $V_\phi$
\STATE Collect warm-up reward vectors $\{\bm{r}_t\}$ by random policy and solve \eqref{eq:empirical_dual} to obtain $\widehat{\blam}(\bm{c})$
  \FOR{each episode}
    \STATE Construct prompt $d(\calF)$ and compute pool embedding $\bm{e} \leftarrow \LLM(d(\calF))$
    \STATE $\bz_0 \leftarrow \textit{BEG}$
    \FOR{$t = 0, 1, \ldots, L_{\max}-1$}
      \STATE $\bm{h}_t \leftarrow \LSTM(\bz_t)$
      \STATE $\bm{u}_t \leftarrow \FFN_h(\bm{h}_t) \odot \FFN_e(\bm{e})$
      \STATE $\bm{q}_t \leftarrow \FFN_u(\bm{u}_t)$
      \STATE Sample $a_t \sim \pi_\theta(\cdot \mid \bm{q}_t)$ with invalid action masking
      \STATE $\bz_{t+1} \leftarrow [\bz_t,\, a_t]$
      \IF{$a_t = \textit{SEP}$ \textbf{or} $|\bz_{t+1}| = L_{\max}$}
        \STATE Parse $f_{\mathrm{new}} \leftarrow \mathrm{parse}(\bz_{t+1})$
        \IF{$f_{\mathrm{new}}$ is valid}
          \STATE Form $\overline{\calF} \leftarrow \calF \cup \{f_{\mathrm{new}}\}$ and refit the linear model on $\overline{\calF}$
          \STATE Update $\calF$ and $\widehat{\bbeta}$ by keeping at most $P$ alphas according to their fitted contributions
          \STATE Compute $\bm{r}_t \leftarrow (\IC_t, \RRE_t, \PFS_t, \mathrm{DH}_t)^\top$
        \ELSE
          \STATE $\bm{r}_t \leftarrow \bm{-1}$
        \ENDIF
        \STATE $r_t^\ast \leftarrow \widehat{\blam}(\bm{c})^\top \bm{r}_t$
        \STATE Store the terminal transition in $\mathcal{B}$
        \STATE \textbf{break}
      \ELSE
        \STATE Store the transition in $\mathcal{B}$ with $\bm{r}_t \leftarrow \bm{0}$ and $r_t^\ast \leftarrow 0$
      \ENDIF

      \IF{$n \,\,\, \mathrm{mod} \,\,\, \overline{n} = 0$}
      \STATE Update $\theta$ and $\phi$ via PPO using a batch of transitions in $\mathcal{B}$
      \STATE Re-solve \eqref{eq:empirical_dual} on recent reward vectors to update $\widehat{\blam}(\bm{c})$
      \ENDIF
      
    \STATE $n \leftarrow n+1$  

    \ENDFOR
    \IF{$n > n^\ast$}
       \STATE \textbf{break}
    \ENDIF
  \ENDFOR
\RETURN $\calF$ and $\widehat{\bbeta}$
\end{algorithmic}
\end{algorithm}

\section{Details About Baseline Methods}\label{sec:baseline}

This section supplements the summary of baseline methods in \cref{sec:experiments}. Unless otherwise noted, all formulaic-alpha discovery baselines are evaluated through the same downstream linear mega-alpha construction, so the comparison focuses on the quality of the selected alpha pool rather than on the complexity of the combiner.

Alpha158 serves as a handcrafted formulaic benchmark. We fix the candidate pool to the 158 predefined alphas provided by Qlib \citep{qlib} and fit a linear model to combine them into the final mega-alpha. In contrast, MLP is a non-formulaic end-to-end predictor that maps input directly to future returns in a data-driven approach. This pair of baselines allows us to compare AlphaPareto with both a standard handcrafted alpha pool and a representative black-box predictor.

The GP baseline follows the conventional genetic-programming paradigm. Candidate expressions are evolved one by one with IC as the fitness measure. We then construct the final alpha pool from the top-$P$ discovered expressions. This baseline represents the classical formulaic-alpha search pipeline that does not directly optimize pool-level synergy during generation.

AlphaAgent and R\&D-Agent-Quant are representative LLM-based alpha discovery methods. Specifically, AlphaAgent uses an LLM to generate market hypotheses and candidate formulas directly, whereas R\&D-Agent-Quant separates this workflow into a research stage for hypothesis generation and a development stage for implementation and backtesting. In both baselines, the LLM acts as a generator of candidate alphas rather than as a frozen semantic encoder inside an RL loop. In particular, we make several adjustments to these baselines to ensure a fair comparison with AlphaPareto. For AlphaAgent, we modify the prompt so that each run generates $10$ candidate alphas per iteration over $5$ iterations. For R\&D-Agent-Quant, we make three changes to the default alpha generation loop: (i) we fix the number of factors per iteration at exactly $10$, instead of the default range of $1$--$5$, by modifying both the prompt template and the proposal module; (ii) we increase the number of outer loops from $5$ to $10$; and (iii) we replace the default free-form LaTeX expression format with the same structured expression syntax used in our method. This syntax covers the full operator set shared across all baselines and allows the generated factors to be parsed and evaluated directly, without requiring LLM-based reverse translation. All other implementation details of AlphaAgent and R\&D-Agent-Quant follow their original repository defaults.

AlphaGen, AlphaForge and AlphaQCM are the closest RL-based competitors. AlphaGen formulates formulaic alpha discovery as an MDP and uses PPO to search for a synergistic alpha pool. AlphaForge modifies the static combination model to a dynamic one. AlphaQCM extends this line by using distributional RL and a variance-based exploration bonus to mitigate the non-stationarity highlighted in our related-work discussion. These methods optimize a scalar reward tied primarily to predictive performance and do not explicitly encode the semantics of the current alpha pool. Their comparison with AlphaPareto therefore isolates the value of pool-aware state augmentation and Pareto-regularized multi-objective optimization.

\section{Hyperparameters}\label{sec:hyper}

\subsection{Network Hyperparameters}\label{sec:network_hyper}

As in \cite{alphagen}, we use a two-layer LSTM for $\LSTM(\cdot)$ in \eqref{LSTM_head}, with hidden dimension $128$ and dropout rate $0.1$. The LLM embedding dimension $d_e$ depends on the chosen model. For example, $d_e = 2560$ for theQwen-Embedding-4B model; see Table \ref{tab:llm_dim} for more details. The hidden dimension of $\FFN_e$ and $\FFN_h$ is set to $256$, and $\FFN_u$ further maps the fused representation $\bm{u}_t$ to a $128$-dimensional embedding. See \eqref{eq:ffns} for details.

\begin{table}[!h]
    \centering
    \setlength{\tabcolsep}{5mm}
    \caption{Embedding dimension of LLM candidates.}
    \begin{tabular}{l|r}
    \toprule
    Chosen model & Embedding dimension $d_e$ \\
    \midrule
    DeepSeek-1.5B & 1536 \\
    Qwen-Embedding-4B & 2560 \\
    DeepSeek-7B & 3584 \\
    DeepSeek-14B & 5120 \\
    DeepSeek-32B & 5120 \\
    \bottomrule
    \end{tabular}\label{tab:llm_dim}
\end{table}

\subsection{RL Hyperparameters}

Besides the network-related hyperparameters, some additional hyperparameters that our RL algorithm inherits are listed in Table \ref{tab:hyper}.

\begin{table}[!h]
    \centering
    \setlength{\tabcolsep}{4.1mm}{
    \caption{Additional hyperparameters.}
    \begin{tabular}{l|r}
    \toprule
    Hyperparameter & Values \\
    \midrule
    Instruments & CSI300 / CSI800 / Market \\
    $P$ (Alpha Pool Size) & 10, 20 \\
    Leading days & 20 \\
    Batch size & 128 \\
    Optimizer & Adam \\
    Learning rate & $5\times10^{-4}$ \\
    Max steps / tolerance & 10,000 / 500 \\
    $\ell_1$ regularization for combination model $\alpha$ & $5\times10^{-3}$ \\
    Lambda solver lr / steps / retries & 0.01 / 500 / 1 \\
    Metrics window size / step & 252 / 5 \\
    Max expression length & 15 \\
    $c_{\IC, CSI300}$ & 0.090 \\
    $c_{\IC, CSI800}$ & 0.085 \\
    $c_{\IC, Market}$ & 0.250 \\
    $c_{\RRE}$ & 0.95 \\
    $c_{\PFS}$ & 0.93 \\
    $c_{\mathrm{DH}}$ & 0.80 \\
    Total step & 250,000 ($P = 10$),\\
    &300,000 ($P = 20$),\\
    \bottomrule
    \end{tabular}\label{tab:hyper}
    }
\end{table}

\section{Paired Significance Tests}
\label{sec:paired_tests}

In addition to comparing means and standard deviations, we further compare AlphaPareto with each baseline in Table~\ref{tab:ic_method} using paired $t$-tests across five matched random seeds, where a positive $t$-statistic favors AlphaPareto. 

Table~\ref{tab:paired_tests} reports the $t$-statistics and their $p$-values. At $10\%$ significance level, AlphaPareto achieves a significantly higher mean of IC than every baseline on CSI300 and CSI800, and than all baselines except AlphaQCM on Market. The comparison with AlphaQCM on Market has a positive estimated difference but is not statistically significant ($t=0.65$, $p=27.60\%$). Overall, the differences favor AlphaPareto and are significant in most comparisons, providing evidence that the predictive gains extend beyond seed-level noise under these tests.

\begin{table}[!h]
  \centering
  \color{black}
  \caption{Paired $t$-tests of AlphaPareto against each baseline across five matched seeds. $p$-values are shown in parentheses as percentages, rounded to two decimal places.}
  \setlength{\tabcolsep}{2.5mm}
  \begin{tabular}{ccccccccc}
    \toprule
          & \multicolumn{2}{c}{CSI300} &       & \multicolumn{2}{c}{CSI800} &       & \multicolumn{2}{c}{Market} \\
    \cmidrule{2-3}\cmidrule{5-6}\cmidrule{8-9}
    Method & $t$ & $p$-value &       & $t$ & $p$-value &       & $t$ & $p$-value \\
    \cmidrule{1-3}\cmidrule{5-6}\cmidrule{8-9}
    Alpha158 & 4.55 & \textcolor[rgb]{ .129,  .361,  .596}{(0.52\%)} &       & 2.40 & \textcolor[rgb]{ .129,  .361,  .596}{(3.73\%)} &       & 13.39 & \textcolor[rgb]{ .129,  .361,  .596}{(0.01\%)} \\
    MLP & 2.73 & \textcolor[rgb]{ .129,  .361,  .596}{(2.62\%)} &       & 7.05 & \textcolor[rgb]{ .129,  .361,  .596}{(0.11\%)} &       & 8.00 & \textcolor[rgb]{ .129,  .361,  .596}{(0.07\%)} \\
    GP & 16.11 & \textcolor[rgb]{ .129,  .361,  .596}{(0.00\%)} &       & 9.58 & \textcolor[rgb]{ .129,  .361,  .596}{(0.03\%)} &       & 10.87 & \textcolor[rgb]{ .129,  .361,  .596}{(0.02\%)} \\
    AlphaAgent & 55.88 & \textcolor[rgb]{ .129,  .361,  .596}{(0.00\%)} &       & 25.41 & \textcolor[rgb]{ .129,  .361,  .596}{(0.00\%)} &       & 44.84 & \textcolor[rgb]{ .129,  .361,  .596}{(0.00\%)} \\
    R\&D-Agent-Quant & 6.59 & \textcolor[rgb]{ .129,  .361,  .596}{(0.14\%)} &       & 34.67 & \textcolor[rgb]{ .129,  .361,  .596}{(0.00\%)} &       & 116.93 & \textcolor[rgb]{ .129,  .361,  .596}{(0.00\%)} \\
    AlphaGen & 1.69 & \textcolor[rgb]{ .129,  .361,  .596}{(8.34\%)} &       & 5.08 & \textcolor[rgb]{ .129,  .361,  .596}{(0.35\%)} &       & 14.96 & \textcolor[rgb]{ .129,  .361,  .596}{(0.01\%)} \\
    AlphaQCM & 14.58 & \textcolor[rgb]{ .129,  .361,  .596}{(0.01\%)} &       & 3.67 & \textcolor[rgb]{ .129,  .361,  .596}{(1.07\%)} &       & 0.65 & \textcolor[rgb]{ .129,  .361,  .596}{(27.60\%)} \\
    AlphaForge & 20.95 & \textcolor[rgb]{ .129,  .361,  .596}{(0.00\%)} &       & 2.74 & \textcolor[rgb]{ .129,  .361,  .596}{(2.59\%)} &       & 9.81 & \textcolor[rgb]{ .129,  .361,  .596}{(0.03\%)} \\
    \textbf{AlphaPareto} & - & \textcolor[rgb]{ .129,  .361,  .596}{-} &       & - & \textcolor[rgb]{ .129,  .361,  .596}{-} &       & - & \textcolor[rgb]{ .129,  .361,  .596}{-} \\
    \bottomrule
  \end{tabular}%
  \label{tab:paired_tests}%
\end{table}

\section{Additional Details and Experiments about Portfolio Backtesting}\label{sec:portfolio}

We further evaluate the out-of-sample economic performance of AlphaPareto through portfolio backtesting. Following the trading strategy in \cite{alphagen} and \cite{alphaforge}, we construct a long-only portfolio by selecting the top $50$ stocks ranked by the mega-alpha at each rebalance date. At each rebalance, we partially rotate the existing holdings by trading up to $5$ eligible stocks, exclude stocks subject to price-limit constraints, and execute trades at the close. Transaction costs are set to $15$ basis points per side, with a minimum fee of $5$ CNY. The backtest is conducted on the full market universe with monthly rebalancing, consistent with the 20-day return prediction target. The initial capital is 100 million CNY, and the Shanghai Composite Index (SH.000001) is used as the benchmark.

\begin{figure}[!h]
    \centering
    \includegraphics[width=1.\textwidth]{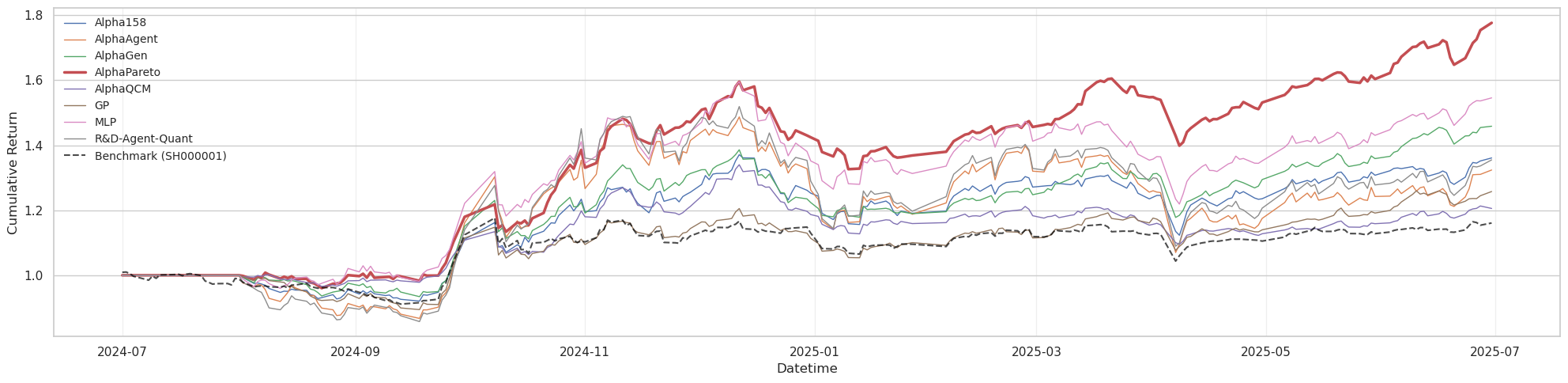}
    \caption{Cumulative returns of monthly portfolios constructed by different methods.}
    \label{fig:portfolio}
\end{figure}

The cumulative returns in \cref{fig:portfolio} provide a consistent visual comparison: AlphaPareto delivers the strongest cumulative return over the test period while maintaining a controlled drawdown profile.

\section{Impact of MORL method}
\label{sec:morl_comparison}

In this paper, we choose MAP, since its value palette specifies interpretable minimum target levels for predictive power, temporal stability, perturbation robustness, and diversity, while the nonnegative scalarization weights are derived from the observed reward distribution. This does not remove subjective preferences, but expresses them as thresholds in the objectives' original units rather than as a fixed weight vector whose meaning changes with objective scale.

While the MORL-only ablation in Table~\ref{tab:ablation_study} shows the benefit of the vector reward design with MAP-based optimization, it does not establish the superiority of MAP over other MORL optimizers. To assess whether MAP's adaptive weighting provides benefits beyond merely combining the four objectives, we consider an equal-weight linear-scalarization baseline that replaces the MAP-derived weights with fixed weights while keeping all other components unchanged. Its scalar reward is the simple average of IC, RRE, PFS, and DH:
\begin{align*}
    r_t^{\mathrm{equal}} = \frac{1}{4}\left(\IC_t + \RRE_t + \PFS_t + \mathrm{DH}_t\right).
\end{align*}
This baseline represents the simplest form of multi-objective scalarization.

On CSI800, the equal-weight baseline achieves a mean IC of $4.92\%$ with a standard deviation of $0.52\%$, compared with $5.70\%$ and $0.87\%$, respectively, for MAP-based AlphaPareto in Table~\ref{tab:ablation_study}. Thus, MAP outperforms this simple alternative in mean IC, although its standard deviation is higher. This comparison supports the benefit of MAP's adaptive weighting beyond merely combining the four objectives. Although these results establish MAP as a strong, well-motivated choice for alpha discovery, more sophisticated MORL optimizers may yield further gains.

\section{Random-Noise Control}
\label{sec:pool_representation_analysis}

To examine whether the LLM encoding acts merely as regularization noise, we replace the LLM-based pool embedding with pool-independent random noise matched in dimension and empirical norm, while keeping all other settings fixed. 

Table~\ref{tab:random_noise_control} compares this control with AlphaPareto and the MORL-only variant, which uses no pool representation. On CSI800, the random-noise control achieves a mean out-of-sample IC of $4.22\%$ with a standard deviation of $1.62\%$, compared with $5.70\%$ and $0.87\%$ for AlphaPareto and $5.54\%$ and $1.03\%$ for the variant without a pool representation. Replacing the LLM embedding with matched noise thus lowers the mean IC and increases its variability. These results suggest that the gain cannot be explained by matched pool-independent noise alone and are consistent with useful pool-dependent semantic information.

\begin{table}[!h]
  \centering
  \caption{Out-of-sample ICs on CSI800 with different pool representations.}
  \setlength{\tabcolsep}{4mm}
  \begin{tabular}{ccc}
    \toprule
          & \multicolumn{2}{c}{CSI800} \\
    \cmidrule{2-3}
    Pool representation & Mean & Std \\
    \cmidrule{1-3}
    None (MORL only) & 5.54\% & \textcolor[rgb]{ .129,  .361,  .596}{(1.03\%)} \\
    Matched noise & 4.22\% & \textcolor[rgb]{ .129,  .361,  .596}{(1.62\%)} \\
    \textbf{LLM embedding (AlphaPareto)} & \pmb{5.70\%} & \textcolor[rgb]{ .129,  .361,  .596}{\pmb{(0.87\%)}} \\
    \bottomrule
  \end{tabular}%
  \label{tab:random_noise_control}%
\end{table}

\section{Empirical Results on the S\&P 500 Universe}
\label{sec:sp500}

To assess whether AlphaPareto remains effective beyond Chinese A-shares, we conduct an additional evaluation on U.S. equities using the S\&P 500 stock universe. We change only the stock universe and retain the other experimental settings described in Section~\ref{sec:experiments}. 

Table~\ref{tab:sp500_ic} reports the out-of-sample predictive performances. On S\&P 500, AlphaPareto achieves the highest mean IC among the evaluated methods and the lowest standard deviation among methods with repeated runs. Compared with the strongest baseline, AlphaQCM, it improves mean IC from $7.18\%$ to $7.65\%$, a gain of $0.47$ percentage points, while reducing the standard deviation from $0.84\%$ to $0.13\%$. These results provide additional empirical support for the applicability of AlphaPareto beyond Chinese A-shares under a different market structure.

\begin{table}[!h]
  \centering
  \caption{Out-of-sample ICs of different methods on the S\&P 500 universe.}
  \setlength{\tabcolsep}{6mm}
  \begin{tabular}{ccc}
    \toprule
          & \multicolumn{2}{c}{S\&P 500} \\
    \cmidrule{2-3}
    Method & Mean & Std \\
    \cmidrule{1-3}
    Alpha158 & 3.92\% & \textcolor[rgb]{ .129,  .361,  .596}{-} \\
    MLP & 5.43\% & \textcolor[rgb]{ .129,  .361,  .596}{(0.31\%)} \\
    AlphaAgent & -0.01\% & \textcolor[rgb]{ .129,  .361,  .596}{(0.56\%)} \\
    R\&D-Agent-Quant & 1.29\% & \textcolor[rgb]{ .129,  .361,  .596}{(0.49\%)} \\
    AlphaGen & 7.03\% & \textcolor[rgb]{ .129,  .361,  .596}{(0.30\%)} \\
    AlphaQCM & 7.18\% & \textcolor[rgb]{ .129,  .361,  .596}{(0.84\%)} \\
    AlphaForge & 1.33\% & \textcolor[rgb]{ .129,  .361,  .596}{(2.97\%)} \\
    \textbf{AlphaPareto} & \pmb{7.65\%} & \textcolor[rgb]{ .129,  .361,  .596}{\pmb{(0.13\%)}} \\
    \bottomrule
  \end{tabular}%
  \label{tab:sp500_ic}%
\end{table}

\section{Computational Resources}\label{sec:compute}

All experiments were conducted on internal GPU servers with 66 Intel(R) Xeon(R) Platinum 8470Q vCPUs, 330 GB RAM, and one RTX PRO 6000 GPU with 96 GB video memory. Each AlphaPareto training run used one GPU and 64 CPU workers. Table \ref{tab:compute_resources} summarizes the compute required to reproduce the reported experiments. All stochastic experiments were repeated over five random seeds, as described in the main text.

The full research project also involved preliminary runs for debugging, prompt design, component designs, etc. These exploratory runs are not included in the main experimental tables. They required approximately 4 additional GPU-hours. The final reported results were produced using the settings in Appendix \ref{sec:hyper}.

\begin{table}[!ht]
    \centering
    \caption{Computational resources used for the reported experiments.}
    \label{tab:compute_resources}
    \setlength{\tabcolsep}{4mm}
    \begin{tabular}{lcccc}
        \toprule
        Experiment & Runs & Hardware per run & Time per run & Total compute \\
        \midrule
        Main comparison & 195 & GPU & 8h & 1560h \\
        LLM scaling study & 150 & GPU & 8h & 1200h \\
        Ablation study & 360 & GPU & 7h & 2520h \\
        \bottomrule
    \end{tabular}
\end{table}



\end{document}